\documentclass[letterpaper]{article} 
\usepackage{_sty/aaai2027}  
\usepackage[hyphens]{url}  
\usepackage{graphicx} 
\usepackage{natbib}  
\usepackage{caption} 
\usepackage{algorithm}
\usepackage{algorithmic}

\usepackage{newfloat}
\usepackage{listings}
\DeclareCaptionStyle{ruled}{labelfont=normalfont,labelsep=colon,strut=off} 
\floatstyle{ruled}
\newfloat{listing}{tb}{lst}{}
\floatname{listing}{Listing}

\usepackage{booktabs}

\usepackage{multirow}
\usepackage{amsmath}
\usepackage{adjustbox}
\usepackage[table]{xcolor}

\title{When and Where to Look: Adaptive Visual Evidence Scheduling for Efficient Long Video Understanding}
\author {
    Ke Li\textsuperscript{\rm 1}\equalcontrib,
    Jiayu Chen\textsuperscript{\rm 2}\equalcontrib,
    Maoliang Li\textsuperscript{\rm 2},
    Zihao Zheng\textsuperscript{\rm 2}, 
    Hailong Zou\textsuperscript{\rm 2}, \\
    Hengyi Zhang\textsuperscript{\rm 2},
    Xuanzhe Liu\textsuperscript{\rm 2},
    Xiang Chen\textsuperscript{\rm 2}\corresponding
}
\affiliations {
    \textsuperscript{\rm 1}School of Electronics Engineering and Computer Science, Peking University\\
    \textsuperscript{\rm 2}School of Computer Science, Peking University\\
}

\nocopyright

\begin{document}

\maketitle

\begin{abstract}
    Efficient long-video understanding requires vision--language models (VLMs) to reason over a small number of frames selected as sparse visual evidence. 
        Existing relevance-based methods rely on static one-shot selection with fixed frame budgets and candidate pools, while agent-based schedulers achieve adaptivity through costly multi-round reasoning and interactive search. 
    We propose EcoFrame, a training-free framework for low-overhead query-adaptive visual evidence scheduling. 
        EcoFrame leverages the VLM's inference feedback to determine when to increase the frame budget and where to search for additional candidate evidence.      
            Specifically, entropy-gated budget scheduling uses output uncertainty to stop early when the current evidence is sufficient or progressively expand the frame budget otherwise. 
            Meanwhile, attention-guided candidate proposal converts frame-level attention into a temporal prior, enabling dense local search in informative regions while preserving global coverage when attention is diffuse. 
    Experiments on Video-MME, LongVideoBench, and MLVU demonstrate that EcoFrame achieves a better accuracy--efficiency trade-off across multiple VLM backbones. On Qwen2.5-VL, EcoFrame achieves an average accuracy of 64.4, surpassing BOLT at 63.5, while providing a $1.85\times$ speedup over AKS and BOLT. Compared with the agent-based A.I.R., EcoFrame maintains comparable accuracy with up to a $13.5\times$ inference speedup. Code will be available at https://github.com/AK-DREAM/EcoFrame. 
\end{abstract}

\section{Introduction}
\label{sec:intr}

Long-video understanding has become an important capability of modern vision--language models (VLMs), supporting applications such as video question answering and video retrieval. 
    However, directly processing an entire long video remains computationally prohibitive, as a video may contain tens of thousands of frames, far exceeding the context window of VLMs. 
In practice, only a small subset of frames can be provided as sparse visual evidence. Consequently, selecting informative evidence under a limited frame budget is critical to both inference efficiency and answer accuracy.

\begin{figure}[t]
    \centering
    \includegraphics[width=\columnwidth]{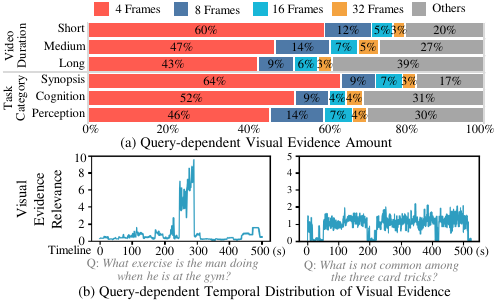}
    \caption{
    Different queries require different amounts and temporal distributions of visual evidence.
    }
    \label{fig:1}
\end{figure}

A straightforward solution is uniform sampling, which selects frames at fixed temporal intervals.
    However, this \textbf{query-agnostic strategy} overlooks a fundamental property of long-video understanding: different queries require different amounts and temporal distributions of visual evidence, as shown in Fig.~\ref{fig:1}. 
A simple query may be answered using only a few frames, whereas a difficult query may require evidence distributed across multiple temporal regions. Therefore, \textit{effective frame selection should adapt both the frame budget and the searched temporal regions to each query.}

\begin{figure}[t]
  \centering
  \includegraphics[width=\columnwidth]{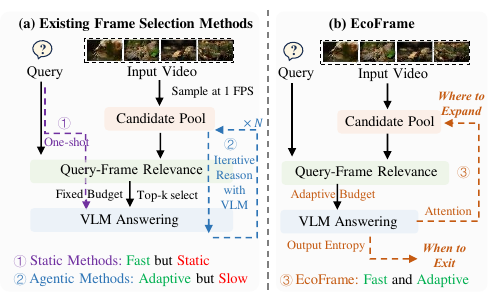}
  \caption{\textbf{Frame selection paradigms.}
Unlike static methods with fixed budgets and candidate pools or agentic methods with costly iterative reasoning, EcoFrame uses VLM internal signals for efficient visual evidence scheduling.}
  \label{fig:2}
\end{figure}

Existing \textbf{query-aware frame selection methods}~\cite{liu2025bolt,zou2026air} can be divided into static and agentic strategies, as illustrated in Fig.~\ref{fig:2}. \textit{Static methods} typically construct a candidate pool before inference, estimate query--frame relevance using a pretrained vision--language encoder such as CLIP, and select a fixed number of frames in a single step. Although efficient, they cannot adjust the frame budget or candidate pool according to the evidence demand of each query. \textit{Agentic methods}, often implemented through agents, improve adaptivity by iteratively reasoning, verifying candidate frames, and searching for frames. However, their reliance on multiple rounds of VLM reasoning and iterative search introduces considerable scheduling overhead.

The prior methods reveal a fundamental trade-off: static selection is efficient but insufficiently adaptive, whereas agentic selection achieves adaptivity at a high reasoning cost. This raises \textbf{the central question of this work}: \emph{how can a VLM adapt its visual evidence to each query without expensive iterative reasoning?} Addressing this problem requires resolving two coupled challenges.
\textbf{Challenge 1: }\textit{Evidence Sufficiency Estimation.}
The required amount of evidence varies across queries and is unknown in advance. A fixed frame budget may be insufficient for some queries yet redundant for others. The model must therefore assess whether the current evidence is sufficient and decide \emph{when to stop or expand} the frame budget.
\textbf{Challenge 2: }\textit{Targeted Temporal Search.}
When the current evidence is insufficient, the model must determine \emph{where to search} for additional frames. A dense candidate pool incurs substantial encoding and relevance-computation costs, whereas a small fixed pool may miss temporally localized evidence. The challenge is to expand the candidate pool toward promising regions without exhaustive processing or costly agent-based search.

To address these challenges with minimal overhead, we analyze two internal signals available during VLM inference: output entropy for answer uncertainty and frame-level attention for temporal focus. This analysis yields two observations.
\textbf{Observation 1: }\textit{Output entropy can indicate evidence sufficiency.}
Under a limited frame budget, insufficient evidence generally leads to higher output entropy. Entropy can therefore guide \emph{when to exit or expand} the frame budget.
\textbf{Observation 2: }\textit{Frame-level attention can indicate where to search.}
Frame-level attention correlates with promising temporal regions for candidate expansion: concentrated attention favors denser local search, whereas diffuse attention calls for broader temporal coverage. Attention can therefore guide \emph{where to expand} the candidate pool.

Motivated by these observations, we propose \textbf{EcoFrame}, a training-free framework for low-overhead query-adaptive visual evidence scheduling. EcoFrame starts from a sparsely sampled candidate pool, selecting a set of relevant frames for a low-budget answer attempt. Instead of employing a separate agent to make scheduling decisions, it directly converts the VLM's output entropy and frame-level attention into explicit scheduling signals during inference.
\textbf{Based on Observation 1}, EcoFrame introduces \emph{entropy-gated budget scheduling}: if the output entropy falls below a round-dependent threshold, the current evidence is considered sufficient and scheduling terminates early; otherwise, the frame budget is progressively expanded.
\textbf{Based on Observation 2}, EcoFrame introduces \emph{attention-guided candidate proposal}, which propagates frame-level attention into a temporal-cell prior to enable denser search in high-attention regions while maintaining global coverage when attention is diffuse. Input frames are reselected from the expanded candidate pool according to query--frame relevance and temporal coverage, forming coarse-to-fine evidence scheduling.

Extensive experiments on LongVideoBench, Video-MME, and MLVU demonstrate that EcoFrame achieves a favorable accuracy--efficiency trade-off across these benchmarks and generalizes across three distinct VLM backbones, including LLaVA-OneVision, Qwen2.5-VL, and InternVL-3. On Qwen2.5-VL, EcoFrame achieves an average accuracy of 64.4, outperforming the best relevance-based method BOLT at 63.5 while providing a $1.85\times$ speedup over AKS and BOLT. Compared with the agent-based method A.I.R., EcoFrame maintains comparable accuracy while reducing inference latency by up to $13.5\times$.

\section{Preliminaries}

\paragraph{Relevance-Based Frame Selection. }
Long-video question answering relies on frame selection to compress a video into a compact set of visual evidence. Formally, given a $N$-frame video $\mathcal{V}=\{f_i\}_{i=1}^{N}$ and a query $Q$, frame selection chooses an input frame set $\mathcal{S}=\{f_{x_i}\}_{i=1}^{K}\subset\mathcal{V}$, where $K\ll N$, for a downstream VLM to predict $\hat{y}=\mathrm{VLM}(\mathcal{S},Q)$. The objective is to select sufficient evidence for accurate answering while minimizing the frame budget $K$.

Uniform sampling selects $K$ frames at fixed intervals and is query-agnostic. Static relevance-based methods instead construct a candidate frame pool $\mathcal{P}$ sampled at a fixed frame rate. They then use a vision--language encoder such as CLIP~\cite{radford2021clip} to compute a query--frame relevance score $s_i=\operatorname{sim}(E_v(f_i),E_t(Q))$ for each $f_i\in \mathcal{P}$, and select $K$ frames based on the relevance score. Although query-aware, these methods often fix both $\mathcal{P}$ and $K$ before inference. A dense pool increases encoding and scoring costs, whereas a sparse pool may miss localized evidence; similarly, a fixed frame budget cannot adapt to evidence sufficiency.

\paragraph{VLM Output Entropy.}
Output entropy characterizes the concentration of the VLM's predictive distribution.
Given an input frame set $\mathcal{S}$ and a query $Q$, the VLM generates an $L$-token answer with predictive distribution $p_{i}(v)$ at position $i$. Let $\mathcal{V}$ denotes the possible token vocabulary. The token entropy and output entropy are computed as
\begin{equation}
\label{eq:entropy}
H_{i}=-\sum_{v\in\mathcal{V}}p_{i}(v)\log p_{i}(v),\quad e_{out}=\frac{1}{L}\sum_{i=1}^{L}H_{i}.
\end{equation}

A lower $e_{\mathrm{out}}$ corresponds to a more concentrated distribution and higher predictive certainty, whereas a higher value indicates a more diffuse distribution and greater uncertainty.

\paragraph{Frame-Level Attention Score.}
During VLM inference, self-attention reflects the model's focus on visual tokens. Following prior work~\cite{endo2025feather,li2026dytok}, we recompute the attention from the last textual query token to all visual tokens before positional embeddings are applied. Let $\mathbf{q}_{\ell,h}$ and $\mathbf{K}_{\ell,h}^{\mathrm{Vis}}$ denote the query vector and visual-token key matrix at head $h$ of layer $\ell$. Given a set of selected deep layers $\mathcal{L}$ and number of attention heads $H$, the token-level attention vector $\mathbf{w}$ is computed as
\begin{equation}
\label{eq:attention}
\mathbf{w}_{\ell,h} = \operatorname{softmax} \left( \frac{\mathbf{q}_{\ell,h}^{\top}\mathbf{K}_{\ell,h}^{\mathrm{Vis}}}{\sqrt{D}} \right), \ \mathbf{w} = \frac{1}{|\mathcal{L}|} \sum_{\ell\in\mathcal{L}} \frac{1}{H} \sum_{h=1}^{H} \mathbf{w}_{\ell,h},
\end{equation}
where $D$ is the head dimension. The frame-level attention score $a_i$ is computed by averaging the entries of $\mathbf{w}$ over the visual tokens of frame $f_i$, representing the attention assigned by the VLM to each input frame during answer generation.
This attention is computed only between a few pairs of tokens in selected layers, adding little overhead while remaining compatible with accelerators such as FlashAttention. 
\section{Motivation and Analysis}
\label{sec:analysis}

EcoFrame is motivated by a simple question: can feedback from low-budget VLM inference reveal \emph{when} additional visual evidence is needed and \emph{where} it should be sought? We study these two decisions through controlled analyses on Video-MME using LLaVA-OneVision and Qwen2.5-VL.

\begin{figure}[t]
    \centering
    \includegraphics[width=\columnwidth]{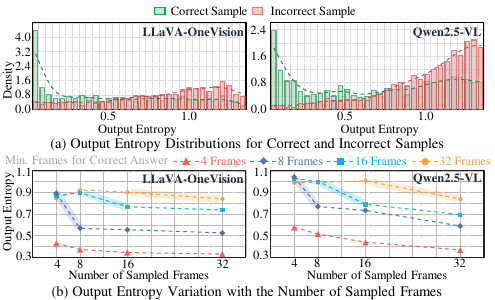}
    \caption{\textbf{Output entropy reflects evidence sufficiency.} Correct answers exhibit lower entropy than incorrect ones, with entropy sharply decreasing when the frame budget reaches the minimum required for a correct answer.
}
    \label{fig:3}
\end{figure}

\paragraph{Observation 1: Output entropy can indicate evidence sufficiency.}
To investigate whether output uncertainty is related to evidence sufficiency, we first compare the output-entropy distributions of correct and incorrect answers under a limited frame budget of 8. As shown in Fig.~\ref{fig:3}(a), correct answers from both VLMs are concentrated in the low-entropy region, whereas incorrect answers occur more frequently at medium or high entropy. This suggests that when the output entropy is low, indicating low model uncertainty, the model is more likely to have already acquired sufficient visual evidence to answer the question correctly.

We further examine how entropy changes as more frames are introduced. Specifically, for each video--query pair, we evaluate frame budgets $\mathcal{B}={4,8,16,32}$ and define $B^\star$ as the smallest budget yielding a correct answer. We group queries by $B^\star$ and track entropy across budgets. As shown in Fig.~\ref{fig:3}(b), for groups with $B^\star>4$, entropy decreases markedly when the budget reaches $B^\star$, i.e., when the model first produces a correct answer after receiving more frames. This pattern is consistent across VLM backbones. Overall, the results reveal that insufficient evidence is generally associated with higher output entropy, while acquiring sufficient evidence coincides with an entropy reduction.

However, absolute entropy levels vary across query groups: even after answering correctly, queries requiring larger $B^\star$ retain higher entropy than those resolved with fewer frames. Output entropy therefore indicates evidence sufficiency but is not a universal correctness certificate. It should be interpreted together with the scheduling round to determine \emph{when to stop or expand} the frame budget.

\begin{figure}[t]
    \centering
    \includegraphics[width=\columnwidth]{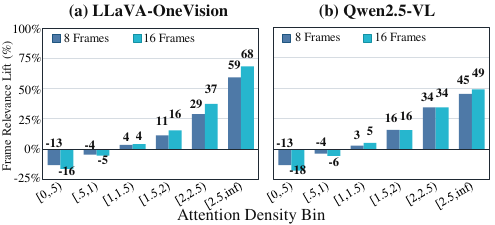}
    \caption{\textbf{The correlation between frame-level attention and temporal evidence.}
    Frame-relevance lift across attention-density bins under 8- and 16-frame inference on (a) LLaVA-OneVision and (b) Qwen2.5-VL. Higher-density bins correspond to more concentrated frame attention.}
    \label{fig:4}
\end{figure}

\paragraph{Observation 2: Frame-level attention can indicate where to search.}
When evidence is insufficient, the problem is where to search for additional frames. To study this, we use the CLIP similarity score between the textual query and all video frames as a proxy for temporal evidence relevance, and compute the frame-level attention score of the input frames to the VLM following Eq.~\ref{eq:attention}. We then group these frames by attention density and compute the relevance lift of their corresponding temporal regions partitioned by adjacent frame midpoints over the video-wide baseline.

Fig.~\ref{fig:4} shows a consistent positive relationship between attention concentration and nearby frame relevance under both 8- and 16-frame inference. Temporal regions around diffuse-attention frames with attention density close to 1 exhibit no significant change in relevance. In contrast, the most concentrated attention bin achieves relevance lifts of $59\%$ and $68\%$ on LLaVA-OneVision and $45\%$ and $49\%$ on Qwen2.5-VL under 8- and 16-frame inference, respectively. The rapid increase across attention-density bins indicates that frames with highly concentrated attention are more likely to lie near query-relevant temporal regions.

This association implies a conditional search prior. Concentrated attention suggests that the VLM may have localized a promising region, favoring local search around high-attention frames. Diffuse attention indicates either that no reliable region has yet emerged or that the required evidence is broadly distributed, favoring global coverage.

\begin{figure*}[ht]
  \centering
  \includegraphics[width=\textwidth,height=2.5in,
  keepaspectratio]{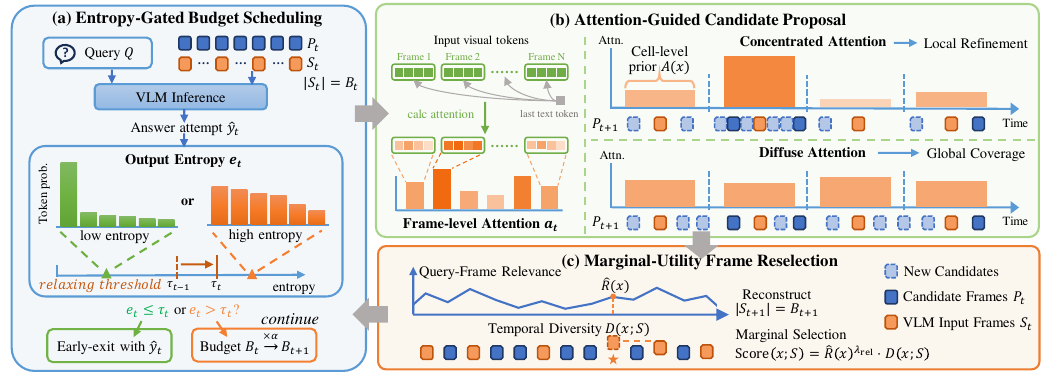}
  \caption{Overview of EcoFrame. Starting from a low-budget input frame set, the target VLM produces an answer, output entropy, and frame-level attention. Entropy controls whether the procedure exits or increases the next-round budget; attention guides candidate pool expansion; relevance and temporal coverage then reconstruct the next input frame set. }
  \label{fig:overview}
\end{figure*}

\section{Method}
\label{sec:method}

\subsection{Overview}
\label{sec:overview}

Given a video $V=\{f_i\}_{i=1}^{N}$ and a query $Q$, EcoFrame performs query-adaptive visual evidence scheduling over a sequence of frame budgets $B_1<\cdots<B_T\leq B_{\max}$. At round $t$, it maintains two frame sets with different roles: a candidate frame pool $P_t$, which stores every frame explored so far together with a cached encoder relevance score, and an input frame set $S_t\subseteq P_t$, which contains exactly $B_t$ frames and is provided to the target VLM. Separating these sets allows the search space to grow incrementally without forcing every explored candidate into the expensive VLM.

Figure~\ref{fig:overview} summarizes the scheduling method. The VLM first answers the query using $S_t$, producing answer $\hat y_t$, output entropy $e_t$, and frame-level attention scores $\mathbf{a}_t$. If $e_t$ passes the entropy gate, EcoFrame returns $\hat y_t$ immediately. Otherwise, it increases the frame budget to $B_{t+1}$, uses $\mathbf{a}_t$ as a temporal prior to expand $P_t$ into $P_{t+1}$, and reconstructs $S_{t+1}$ from the enlarged pool based on query relevance and coverage. This closed loop progressively refines the visual evidence by reusing signals from the target VLM's own answer attempt, avoiding additional reasoning or verification.

\subsection{Entropy-Gated Budget Scheduling}\label{sec}

\textbf{Based on Observation 1}, output entropy can serve as a low-overhead evidence-sufficiency signal, guiding \emph{when to stop or expand} the frame budget. Specifically, at the $t$-th inference round, we compute the output entropy $e_t$ over the generated answer following Eq.~\ref{eq:entropy}. A lower $e_t$ indicates a more concentrated predictive distribution, suggesting that the current visual evidence is more likely to be sufficient.

To translate this sufficiency signal into a bounded scheduling procedure, EcoFrame compares $e_t$ with a round-dependent threshold $\tau_t$.  If $e_t<\tau_t$, the current evidence is considered sufficient, and scheduling terminates early with answer $\hat{y}_t$. If $B_t=B_{\max}$, EcoFrame also terminates to ensure a bounded procedure. Otherwise, it progressively expands the frame budget as $B_{t+1}=\min(\lceil\alpha B_t\rceil,B_{\max})$, where $\alpha>1$. This geometric schedule enables repeated sufficiency estimation at intermediate budgets while reaching $B_{\max}$ in only logarithmically many rounds.

\paragraph{Relaxing Threshold.} A global threshold cannot accommodate the query-dependent entropy levels observed in Fig.~\ref{fig:3}(b): queries requiring more frames often retain higher entropy even after obtaining sufficient evidence. We therefore use a relaxing threshold $\tau_t=\tau_1+(t-1)\Delta_{\tau}$ with $\Delta_{\tau}\geq0$. Early rounds apply a stricter threshold to prevent premature stopping, whereas later rounds tolerate higher uncertainty of difficult queries, avoiding unnecessary expansion to $B_{\max}$.

\subsection{Attention-Guided Candidate Proposal}
\label{sec:candidate_proposal}

\textbf{Based on Observation 2}, frame-level attention provides a low-overhead temporal search prior, guiding \emph{where to search} when expanding the candidate pool. Following Eq.~\ref{eq:attention}, we compute the token-level attention vector $\mathbf{w}_t$ during the $t$-th inference round. To convert $\mathbf{w}_t$ into a 
frame-level prior that guides candidate proposal, we average $\mathbf{w}_t$ over the tokens of each frame $f_{t,i}\in S_t$ to compute its attention score: 
\begin{equation}
    a_{t,i}=\frac{1}{|\mathcal{I}_{t,i}|}\sum_{j\in\mathcal{I}_{t,i}} \mathbf{w}_{t,j},
\end{equation}
where $\mathcal{I}_{t,i}$ indexes the visual tokens of frame $f_{t,i}$.

Yet, these frame scores exist only on the sparse frame set $S_t$ and cannot directly guide full-video search. We therefore propagate them across the timeline using temporal cells. After sorting the input frames by timestamp, adjacent midpoints define a one-dimensional partition $\{\mathcal{C}_{t,i}\}_{i=1}^{B_t}$. Each unobserved location $x\in\mathcal{C}_{t,i}$ inherits the cell-level prior $A_t(x)=a_{t,i}$, which serves as a timeline-wide search signal.

When the frame budget expands from $B_t$ to $B_{t+1}$, EcoFrame first determines the number of new candidates as $M_{t+1}=\lceil m\cdot g(N)\cdot \Delta B_{t+1}\rceil$, where $\Delta B_{t+1}=B_{t+1}-B_t$, $m$ is the candidate expansion ratio, and $g(N)$ is a capped length-aware factor. However, using the propagated prior alone may cluster proposals within high-attention regions. To balance relevance and temporal coverage, an unobserved frame $x$ is scored against the current pool $P$ by
\begin{align}
D(x;P)
&=\min_{y\in P}\frac{|x-y|}{N},\\
\operatorname{Score}_{\mathrm{cand}}(x;P)
 &=
A_t(x)^{\lambda_{\mathrm{attn}}}\cdot 
  D(x;P),
\label{eq:proposal_score}
\end{align}
where $\lambda_{\mathrm{attn}}$ controls the sharpness of the attention prior. The attention prior term favors promising regions, while the distance term $D(x;P)$ discourages redundant proposals near existing frames.
Since each proposed frame changes the remaining temporal distances, EcoFrame starts from $P:=P_t$ and repeatedly adds the highest-scoring frame and updates the distances until $M_{t+1}$ candidates are proposed. The query--frame relevance of these newly proposed candidates are then computed and cached to form $P_{t+1}$. 

\paragraph{Analysis.} The multiplicative proposal score yields the desired switch between exploitation and exploration. When attention is concentrated, candidates within high-attention cells receive higher score, inducing dense local search. When attention is diffuse, the cell priors become flatter and the distance term dominates, allocating candidates to poorly covered regions across the video. Intuitively, the number of candidates assigned to each cell approximates
\begin{equation}
\operatorname{Num}_{\mathrm{cand}}(\mathcal{C}_{t,i})
\propto
a_{t,i}^{\lambda_{\mathrm{attn}}} \cdot\operatorname{length}(\mathcal{C}_{t,i}),
\label{eq:cell_allocation}
\end{equation}
so the same mechanism adapts continuously between local refinement and global coverage.

\subsection{Marginal-Utility Frame Reselection}
\label{sec:frame_reselection}

Candidate proposal broadens the inspected region, but the resulting pool generally exceeds the target VLM's input budget. EcoFrame therefore reconstructs $S_{t+1}$ from the full pool $P_{t+1}$. We first normalize the cached query--frame relevance $R(x)=
\operatorname{sim}\!\left(E_v(f_x),E_t(Q)\right)$ of each candidate frame $f_x\in P_{t+1}$ into a $[0,1]$ distribution:
$
\widehat R=\text{softmax}(R / \sigma_R),
\label{eq:normalized_relevance}
$
where $\sigma_R$ is the standard deviation of all $R(x)$. 

Selecting by relevance alone may concentrate frames around a single temporal event. To balance relevance and diversity, we measure the marginal utility of adding candidate $x$ to a partially constructed set $S$ as:
\begin{equation}
\operatorname{Score}_{\mathrm{select}}(x;S)
=\widehat R(x)^{\lambda_{\mathrm{rel}}} \cdot
  D(x;S),
\label{eq:selection_score}    
\end{equation}
where $\lambda_{\mathrm{rel}}$ controls relevance sharpness. Following a similar procedure to candidate pool expansion, EcoFrame initializes $S$ with the most relevant candidate and repeatedly adds the highest-scoring remaining frame until $|S|=B_{t+1}$. Reselecting from the full candidate pool allows newly discovered high-relevance frames to replace weaker earlier frames, while the distance term prevents temporal collapse.

\paragraph{Initialization.} Since no frame-level attention prior is available before the first VLM call, EcoFrame uniformly samples an initial pool $P_1$ of size $M_1=\lceil m\cdot g(N)\cdot B_1\rceil$, caches the query--frame relevance scores, and applies the same procedure to construct the initial evidence set $S_1$.

\paragraph{Bounded Computation.} Although the frame budget grows across rounds, both sources of visual computation remain bounded. The VLM input never exceeds $B_{\max}$, and the geometric schedule bounds cumulative VLM processing by a geometric sum. Since candidate expansion follows the budget increment and $g(N)$ is capped, frame encoding and scoring cost are bounded by $\mathcal{O}(m\,g_{\max}B_{\max})$. Overall, easy queries terminate with a small budget, whereas difficult queries receive additional but bounded evidence acquisition.

\section{Experiments}

\newcommand{\air}[1]{#1}

\begin{table*}[t]
\centering
\setlength{\tabcolsep}{3.6pt}
\renewcommand{\arraystretch}{1.08}
\begin{adjustbox}{max width=\textwidth}
\begin{tabular}{llcccccccccc}
\toprule
\multirow{2}{*}{\textbf{Model}}
& \multirow{2}{*}{\textbf{Method}}
& \multirow{2}{*}{\textbf{\#Frames}}
& \multicolumn{2}{c}{\textbf{Video-MME}}
& \multicolumn{2}{c}{\textbf{LongVideoBench}}
& \multicolumn{2}{c}{\textbf{MLVU}}
& \multicolumn{3}{c}{\textbf{Average}} \\
\cmidrule(lr){4-5}
\cmidrule(lr){6-7}
\cmidrule(lr){8-9}
\cmidrule(lr){10-12}
& &
& \textbf{Acc. $\uparrow$}
& \textbf{Lat. $\downarrow$}
& \textbf{Acc. $\uparrow$}
& \textbf{Lat. $\downarrow$}
& \textbf{Acc. $\uparrow$}
& \textbf{Lat. $\downarrow$}
& \textbf{Acc. $\uparrow$}
& \textbf{Lat. $\downarrow$}
& \textbf{Speedup $\uparrow$} \\
\midrule

\multirow{6}{*}{LLaVA-OV-7B}
& Uniform & 32
& 58.6 & 1.66
& 56.6 & 1.67
& 63.1 & 1.68
& 59.4 & 1.67 & -- \\
\cmidrule(lr){2-12}

& AKS & 32
& 59.2 & 4.50
& 57.5 & 3.67
& 67.6 & 3.46
& 61.4 & 3.88 & 1.0$\times$ \\

& BOLT & 32
& 60.4 & 4.50
& 58.3 & 3.67
& 68.0 & 3.47
& 62.2 & 3.88 & 1.0$\times$ \\

& FOCUS & 32
& 58.0 & 2.92
& 59.2 & 2.56
& 65.0 & 2.48
& 60.7 & 2.65 & 1.46$\times$ \\

& A.I.R.$^{\dagger}$ & 22.3
& \textbf{\air{61.4}} & 28.95
& \textbf{\air{60.7}} & 32.68
& \underline{\air{69.3}} & 25.98
& \textbf{\air{63.8}} & 29.20 & 0.13$\times$ \\

\rowcolor{gray!10}
& \textbf{EcoFrame} & 19.1
& \underline{61.0} & \textbf{2.45}
& \underline{59.9} & \textbf{2.19}
& \textbf{69.8} & \textbf{1.87}
& \underline{63.5} & \textbf{2.17}
& \textbf{1.79$\times$} \\


\midrule

\multirow{6}{*}{Qwen2.5-VL-7B}
& Uniform & 32
& 60.4 & 1.02
& 58.9 & 1.03
& 59.2 & 0.99
& 59.5 & 1.01 & -- \\
\cmidrule(lr){2-12}

& AKS & 32
& 62.4 & 3.85
& 57.4 & 3.04
& 63.9 & 2.77
& 61.2 & 3.22 & 1.0$\times$ \\

& BOLT & 32
& 62.9 & 3.86
& 59.3 & 3.04
& \underline{68.4} & 2.76
& 63.5 & 3.22 & 1.0$\times$ \\

& FOCUS & 32
& 60.0 & 2.27
& 58.4 & 1.90
& 64.0 & 1.79
& 60.8 & 1.99 & 1.62$\times$ \\

& A.I.R.$^{\dagger}$ & 23.1
& \textbf{\air{65.0}} & 21.03
& \textbf{\air{61.4}} & 25.30
& \air{67.5} & 19.97
& \textbf{\air{64.6}} & 22.10 & 0.15$\times$ \\

\rowcolor{gray!10}
& \textbf{EcoFrame} & 21.3
& \underline{64.4} & \textbf{1.92}
& \underline{60.3} & \textbf{1.81}
& \textbf{68.6} & \textbf{1.49}
& \underline{64.4} & \textbf{1.74}
& \textbf{1.85$\times$} \\


\midrule

\multirow{6}{*}{InternVL-3-8B}
& Uniform & 32
& 64.6 & 1.64
& 57.9 & 1.64
& 67.7 & 1.75
& 63.4 & 1.68 & -- \\
\cmidrule(lr){2-12}

& AKS & 32
& 66.2 & 4.47
& 58.5 & 3.65
& 73.2 & 3.53
& 66.0 & 3.88 & 1.0$\times$ \\

& BOLT & 32
& 66.7 & 4.47
& 60.4 & 3.64
& 73.6 & 3.53
& 66.9 & 3.88 & 1.0$\times$ \\

& FOCUS & 32
& 64.2 & 2.90
& 59.3 & 2.55
& 69.5 & 2.54
& 64.3 & 2.66 & 1.46$\times$ \\

& A.I.R.$^{\dagger}$ & 22.6
& \textbf{\air{68.2}} & 26.74
& \textbf{\air{62.8}} & 31.04
& \textbf{\air{74.5}} & 23.58
& \textbf{\air{68.5}} & 27.12 & 0.14$\times$ \\

\rowcolor{gray!10}
& \textbf{EcoFrame} & 17.4
& \underline{68.0} & \textbf{2.48}
& \underline{61.1} & \textbf{2.20}
& \underline{74.3} & \textbf{1.48}
& \underline{67.8} & \textbf{2.05}
& \textbf{1.89$\times$} \\


\bottomrule
\end{tabular}
\end{adjustbox}

\caption{
Main results across different datasets and VLMs.
We report accuracy (\%) and average end-to-end GPU latency (sec).
\textbf{Bold} and \underline{underlined} values indicate the best and
second-best results, respectively.
$^{\dagger}$ denotes an agent-based method.
}
\label{tab:main_results}
\end{table*}

\begin{table}[t]
\centering
\footnotesize
\begin{adjustbox}{max width=\columnwidth}
\begin{tabular}{lcccc}
\toprule
\multirow{2}{*}{\textbf{Method}}
& \multirow{2}{*}{\textbf{Acc.}}
& \multicolumn{3}{c}{\textbf{Latency (s)}} \\
\cmidrule(lr){3-5}
& & \textbf{Scoring} & \textbf{Inference} & \textbf{End-to-end} \\
\midrule
Uniform & 56.6 & -- & 1.67 & 1.67 \\
BOLT & 58.3 & 2.01 & 1.66 & 3.67 (1.00$\times$) \\
FOCUS & 59.2 & 0.90 & 1.66 & 2.56 (1.43$\times$) \\
\midrule
\textbf{EcoFrame} & \textbf{59.9} & 0.48 & 1.71 & 2.19 (\textbf{1.68$\times$}) \\
\quad  exit $\leq 8$f & -- & 0.13 & 0.38 & 0.51 \\
\quad  exit $\geq 16$f & -- & 0.81 & 2.66 & 3.47 \\
\bottomrule
\end{tabular}
\end{adjustbox}
\caption{
Latency breakdown on LongVideoBench (avg. 12 min) with LLaVA-OV-7B.
We decompose the GPU latency into candidate frame scoring and VLM inference. Queries are grouped by their used frame budgets when exit.
}
\label{tab:latency_breakdown}
\end{table}

\subsection{Experimental Setup}
\paragraph{Datasets and Metrics.} We conduct experiments using the lmms-eval~\cite{zhang2025lmms} framework on three widely adopted long-video QA benchmarks: Video-MME~\cite{fu2025videomme}, LongVideoBench~\cite{wu2024longvideobench}, and MLVU~\cite{zhou2025mlvu}. For each dataset and VLM backbone, we report average accuracy and end-to-end GPU latency. We also report the average number of frames used for final answering, representing the average frame budget.

\paragraph{Baselines and Models.} We compare EcoFrame with a set of representative frame selection methods, including vanilla uniform sampling, static relevance-based selection methods such as AKS, BOLT, and FOCUS~\cite{tang2025aks,liu2025bolt,zhu2026focus}, and an agent-based iterative method A.I.R.~\cite{zou2026air}. 
To assess generalizability across different models, we conduct experiments using three VLMs with diverse architectures: LLaVA-Onevision-7B, Qwen2.5-VL-7B, and InternVL-3-8B~\cite{li2024llava,bai2025qwen25vl,zhu2025internvl3}.

\paragraph{Implementation Details. } For EcoFrame, we initialize the frame budget with $B_1=4$ frames and multiply it by $\alpha=2$ each round. For the early-exit thresholds, we set $\tau_1=0.1$ and $\Delta_\tau=0.2$ for Video-MME and $\tau_1=0.2$ and $\Delta_\tau=0.3$ for LongVideoBench and MLVU to target comparable efficiency regimes. We fix other hyperparameters to $m=4$, $\lambda_{\text{attn}}=0.5$, and $\lambda_{\text{rel}}=1$ across all experiments. To ensure fair comparison, all methods use the same maximum frame budget of 32 and the same CLIP-ViT-L/14 model for query--frame relevance scoring. Relevance-based methods use their default 1 fps candidate pool, and all other hyperparameters follow their official settings. All experiments are conducted on a Linux server with 2 NVIDIA L20 GPUs.

\subsection{Main Results}
Table~\ref{tab:main_results} reports the main results across different datasets and VLM backbones. Overall, EcoFrame achieves a superior accuracy--efficiency trade-off against the baselines. 

Compared with uniform sampling and static relevance-based selection methods, EcoFrame consistently improves average accuracy by up to \textbf{4.9\%} while using only about 17--21 frames in the final round, and achieves up to \textbf{1.89$\times$} speedup over AKS and BOLT. 
This demonstrates the advantage of query-adaptive evidence scheduling compared to one-shot frame selection. 
By adapting the frame budget to the evidence demand of each query, EcoFrame avoids redundant computation for simple queries while allocating sufficient budget and targeted evidence search for harder ones. 

Compared with the agent-based method A.I.R., EcoFrame maintains comparable accuracy while reducing the average latency by up to \textbf{13.5$\times$}. This efficiency gain comes from its lightweight scheduling mechanism: instead of performing explicit multi-step reasoning or costly VLM-based frame verification, EcoFrame reuses the inference-time entropy and attention signals from the target VLM's low-budget answer attempt to decide whether to expand the frame budget and where to search next. As a result, EcoFrame preserves the adaptivity of closed-loop evidence acquisition while substantially reducing scheduling overhead.

\subsection{Efficiency Analysis}

Table~\ref{tab:latency_breakdown} decomposes the end-to-end latency into candidate frame scoring and VLM inference. Other scheduling operations with negligible overhead are omitted. Compared with uniform sampling, EcoFrame introduces a small additional cost but yields a substantial accuracy improvement. Compared with static relevance-based methods, EcoFrame provides a more adaptive computation allocation: simple queries can exit early with less candidate frame scoring and VLM computation, while hard queries use additional but bounded refinement rounds. Meanwhile, the attention pattern from coarse inference rounds guide adaptive evidence search, reducing the need for dense frame scoring and improves accuracy. Overall, by adaptively allocating computation according to the evidence demand of each query, EcoFrame achieves higher accuracy with lower average latency.

\subsection{Ablation Studies}

\paragraph{Module Ablation.} Table~\ref{tab:module_ablation} systematically evaluates the contribution of each proposed module. 
First, entropy-gated budget scheduling achieves a better accuracy--cost trade-off than fixed-budget inference. Compared with always using the maximum 32 frames budget, it substantially reduces latency with only a small accuracy drop, showing that EcoFrame can adapt the frame budget to the evidence demand of each query.
Second, attention-guided candidate proposal is crucial for efficient evidence search. A static fixed-rate pool with same average candidate number suffers from non-adaptability and often misses crucial evidence. In contrast, EcoFrame progressively expands the candidate pool across rounds, and the accuracy gain over the w/o attention prior variant confirms the critical role of attention-guided proposal.
Finally, marginal-utility frame reselection outperforms relevance-only or coverage-only selection by ensuring both representativeness and diversity in the selected frame set.

\begin{table}[t]
\centering
\footnotesize
\begin{adjustbox}{max width=\columnwidth}
\begin{tabular}{lcccc}
\toprule
\multirow{2}{*}{\textbf{Variant}}
& \multicolumn{2}{c}{\textbf{Video-MME}}
& \multicolumn{2}{c}{\textbf{LongVideoBench}} \\
\cmidrule(lr){2-3}
\cmidrule(lr){4-5}
& \textbf{Acc.} & \textbf{Lat.}
& \textbf{Acc.} & \textbf{Lat.} \\
\midrule
Uniform 32 frames
& 58.6 & 1.66
& 56.6 & 1.67 \\
\textbf{EcoFrame}
& 61.0 & 2.45
& 59.9 & 2.19 \\
\midrule
\multicolumn{5}{l}{\textbf{\textit{(A) w/o Entropy-Gated Budget Scheduling}}} \\
\quad Fixed budget = 8f
& 57.6 & 0.79
& 57.1 & 0.78 \\
\quad Fixed budget = 16f
& 59.2 & 1.72
& 58.3 & 1.70 \\
\quad Fixed budget = 32f
& 61.5 & 3.79
& 60.3 & 3.75 \\
\midrule
\multicolumn{5}{l}{\textbf{\textit{(B) w/o Attention-Guided Candidate Proposal}}} \\
\quad Fixed candidate pool
& 59.0 & 2.52
& 58.3 & 2.01 \\
\quad w/o attention prior
& 60.7 & 2.46
& 59.2 & 2.18 \\
\midrule
\multicolumn{5}{l}{\textbf{\textit{(C) w/o Marginal-Utility Frame Reselection}}} \\
\quad Relevance-only
& 60.6 & 2.45
& 58.7 & 2.18 \\
\quad Coverage-only
& 57.6 & 2.45
& 56.8 & 2.19 \\
\bottomrule
\end{tabular}
\end{adjustbox}
\caption{
Ablation of components on Video-MME and LongVideoBench with LLaVA-OV-7B.
}
\label{tab:module_ablation}
\end{table}

\paragraph{Entropy Thresholds.} 

\begin{table}[t]
\centering
\small
\begin{tabular}{lccc}
\toprule
\textbf{\qquad \quad $\tau_1 / \Delta_{\tau}$}
& \textbf{\#Frames}
& \textbf{Acc.}
& \textbf{Latency} \\
\midrule
$\tau_1=0.1,\ \Delta_{\tau}=0.0$
& 25.0
& 61.2
& 2.95 \\
$\tau_1=0.1,\ \Delta_{\tau}=0.2$
& 21.0
& 61.0
& 2.45 \\
$\tau_1=0.1,\ \Delta_{\tau}=0.3$
& 19.1
& 60.6
& 2.21 \\
\midrule
$\tau_1=0.2,\ \Delta_{\tau}=0.0$
& 22.6
& 60.7
& 2.66 \\
$\tau_1=0.2,\ \Delta_{\tau}=0.2$
& 19.3
& 60.4
& 2.24 \\
$\tau_1=0.2,\ \Delta_{\tau}=0.3$
& 17.2
& 60.0
& 1.97 \\
\bottomrule
\end{tabular}
\caption{
Analysis of different early-exit entropy thresholds on Video-MME with LLaVA-OV-7B.
Each row reports the average final frame budget, accuracy, and latency.
}

\label{tab:entropy_thresholds}
\end{table}

Table~\ref{tab:entropy_thresholds} analyzes the impact of different early-exit entropy thresholds.
The results show a consistent trade-off: stricter thresholds allocate more budget to each query, leading to higher accuracy but also higher latency.
In addition, compared with a fixed threshold where $\Delta_{\tau}=0$, progressively relaxing the threshold across rounds achieves a better trade-off, consistent with our observation in Fig.~\ref{fig:3} that harder queries tend to retain higher entropy and benefit from more tolerant later-round thresholds.

\paragraph{Hyperparameter Sensitivity.} 

\begin{figure}[ht]
    \centering
    \includegraphics[width=\columnwidth]{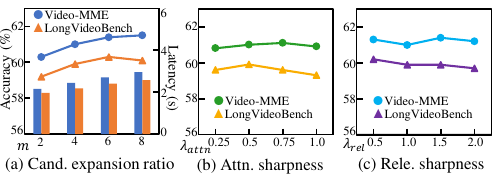}
    \caption{
    Ablation of candidate expansion ratio $m$,
    attention sharpness $\lambda_{\mathrm{attn}}$,
    and relevance sharpness $\lambda_{\mathrm{rel}}$ with LLaVA-OV-7B. 
    Accuracy in lines, latency in bars.
    }
    \label{fig:hyperparam_sweep}
\end{figure}

Figure~\ref{fig:hyperparam_sweep} analyzes the sensitivity of EcoFrame to three key hyperparameters.
Increasing the candidate expansion ratio $m$ enlarges the candidate pool and improves accuracy at first, but the gain saturates while latency continues to increase. EcoFrame is less sensitive to the attention and relevance sharpness parameters $\lambda_{\mathrm{attn}}$ and $\lambda_{\mathrm{rel}}$, showing stable accuracy across a reasonable range without careful tuning. We therefore use moderate default values for all experiments. More results are included in appendix.
\section{Related Work}

\paragraph{Video Understanding VLMs.}
Modern vision--language models (VLMs) have evolved from image-centric visual instruction tuning, exemplified by LLaVA~\cite{liu2023visual}, to video-specialized systems such as Video-LLaVA and Video-ChatGPT~\cite{lin2024videollava,maaz2024videochatgpt}, as well as general-purpose multimodal backbones including LLaVA-OneVision, Qwen2.5-VL, and InternVL3~\citep{li2024llava,bai2025qwen25vl,zhu2025internvl3}. To scale beyond short clips, recent works explores context and memory compression for VLMs~\citep{song2024moviechat,zhang2024longva,shen2025longvu,shu2025videoxl}. 
However, long videos often contain far more frames than can be densely processed under practical computation and memory budgets. Consequently, extracting a compact set of essential visual evidence remains critical to both inference efficiency and answer accuracy.

\paragraph{Frame Selection for Long-Video Understanding.}
To fit long videos within VLM context windows, relevance-based frame selection methods improve over query-agnostic uniform sampling by scoring the query relevance of candidate frames using pretrained encoders such as CLIP~\cite{sun2025mdp3,zhang2025qframe,chen2026wavelet,ma2026gift}. AKS and BOLT balance relevance with coverage or diversity~\cite{tang2025aks,liu2025bolt}, while FOCUS and T* reduce scoring costs by exploring promising temporal regions~\cite{zhu2026focus,ye2025tstar}. Nevertheless, these methods often perform one-shot selection from a pre-constructed candidate pool and cannot adapt the frame budget or candidate search based on query-specific evidence requirements and VLM feedback. Agent-based approaches such as VideoAgent and A.I.R. instead use iterative reasoning and evidence verification to guide subsequent searches, improving adaptivity at considerable computation and latency~\cite{wang2024videoagent,zou2026air,ding2026videozoomer}. 
\section{Conclusion}
We introduce EcoFrame, a training-free framework for low-overhead query-adaptive visual evidence scheduling.
By using output entropy to adapt the frame budget and frame-level attention to guide candidate expansion, EcoFrame achieves a strong accuracy--efficiency trade-off without expensive agent-based reasoning.
Our results demonstrate the value of reusing internal signals produced by the target VLM for lightweight, adaptive evidence scheduling.

\clearpage
\bibliography{_ref/aaai2027}

\clearpage
\appendix
\section{Additional Experimental Results}

In this section, we provide additional experimental results that complements the main paper. All experiments use default hyperparameters and LLaVA-Onevision-7B as the VLM backbone if not specified. 

\subsection{Different Vision--Language Encoders}
Table~\ref{tab:vl_encoder} compares different vision--language encoders for frame scoring: CLIP-ViT-B, CLIP-ViT-L~\cite{radford2021clip}, and SigLIP-so400m~\cite{zhai2023sigmoid}.

EcoFrame consistently outperforms vanilla uniform sampling across all three encoders, demonstrating its robustness to the encoder choice.
Although a smaller encoder like CLIP-ViT-B achieves lower encoding latency, its weaker representations lead to reduced accuracy.
We adopt CLIP-ViT-L as the default encoder for its favorable performance.

\begin{table}[h]
\centering
\small
\begin{tabular}{lcccccc}
\toprule
\multirow{2}{*}{\textbf{Encoder}}
& \multicolumn{2}{c}{\textbf{Video-MME}}
& \multicolumn{2}{c}{\textbf{LVB}}
& \multicolumn{2}{c}{\textbf{MLVU}} \\
\cmidrule(lr){2-3}
\cmidrule(lr){4-5}
\cmidrule(lr){6-7}
& \textbf{Acc.}
& \textbf{Lat.}
& \textbf{Acc.}
& \textbf{Lat.}
& \textbf{Acc.}
& \textbf{Lat.} \\
\midrule
Uniform 
& 58.6 & 1.66
& 56.6 & 1.67
& 63.1 & 1.68 \\
\midrule
CLIP-ViT-B
& 60.4 & 1.93
& 58.9 & 1.79
& 66.8 & 1.51 \\
CLIP-ViT-L
& 61 & 2.45
& \textbf{59.9} & 2.19
& \textbf{69.8} & 1.87 \\
SigLIP-so400m
& \textbf{61.1} & 3.79
& 58.7 & 3.29 
& 69.3 & 2.78 \\
\bottomrule
\end{tabular}
\caption{
Ablation of different vision--language encoders for computing query--frame relevance scores. 
}
\label{tab:vl_encoder}
\end{table}

\subsection{Different Frame Budget Schedules}
Table~\ref{tab:budget_schedule} compares alternative frame budget schedules against our default schedule $B=[4,8,16,32]$.

Reducing the maximum budget from 32 to 16 reduces latency but results in a clear accuracy drop.
Starting from 8 frames prevents early exits at smaller budgets and achieves higher accuracy on Video-MME at the cost of increased latency.
Directly jumping from 4 to 32 frames reduces the number of refinement rounds,
but removes intermediate opportunities to reassess evidence sufficiency and redirect the evidence search, leading to lower accuracy.

\begin{table}[h]
\centering
\small
\begin{tabular}{lcccc}
\toprule
\multirow{2}{*}{\textbf{Schedule}}
& \multicolumn{2}{c}{\textbf{Video-MME}}
& \multicolumn{2}{c}{\textbf{LongVideoBench}} \\
\cmidrule(lr){2-3}
\cmidrule(lr){4-5}
& \textbf{Acc.}
& \textbf{Lat.}
& \textbf{Acc.}
& \textbf{Lat.} \\
\midrule
$B=[4,8,16]$
& 58.7
& 1.37
& 58.4
& 1.31 \\

$B=[8,16,32]$
& \textbf{61.4}
& 2.70
& 59.8
& 2.55 \\

$B=[4,32]$
& 60.8
& 2.26
& 59.4
& 2.13 \\

$B=[4,8,16,32]$
& 61.0
& 2.45
& \textbf{59.9}
& 2.19 \\
\bottomrule
\end{tabular}
\caption{
Ablation of different frame budget schedules. The schedule list $B$ denotes the frame budget of each round. 
}
\label{tab:budget_schedule}
\end{table}

\subsection{Length-aware Candidate Number}
During candidate pool expansion, the candidate number is partially determined by a capped length-aware factor $g(N)$ that grows sublinearly with the video length $N$:

\begin{equation}
\label{eq:length_factor}
    g(N)=\min(\max(\sqrt{N/N_0}, 1), g_{max}),
\end{equation}
where we set $N_0$ to 2 min and $g_{max}=4$. This allows the candidate number to scale mildly to video length,
while avoiding the unbounded scoring cost caused by linear growth.

To validate this design, we compare different length-aware factors in Table~\ref{tab:length_factor}.
A constant factor cannot scale with video length and may miss relevant evidence in longer videos,
while linear growth produces denser candidate pools at the cost of substantial redundant scoring.
In contrast, capped sublinear growth is sufficient when combined with attention-guided proposal,
yielding a better accuracy--efficiency trade-off.

\begin{table}[h]
\centering
\small
\begin{tabular}{lcccc}
\toprule
\multirow{2}{*}{$g(N)$}
& \multicolumn{2}{c}{\textbf{Video-MME}}
& \multicolumn{2}{c}{\textbf{LongVideoBench}} \\
\cmidrule(lr){2-3}
\cmidrule(lr){4-5}
& \textbf{Acc.}
& \textbf{Lat.}
& \textbf{Acc.}
& \textbf{Lat.} \\
\midrule
Constant
& 60.3
& 2.14
& 59.5
& 1.95 \\

Linear
& 60.9
& 4.07
& \textbf{60.1}
& 3.08 \\

Sublinear (Ours)
& \textbf{61.0}
& 2.45
& 59.9
& 2.19 \\
\bottomrule
\end{tabular}
\caption{
Comparison of different length-aware factors for determining the candidate number. Constant ($g(N)=1$), Linear ($g(N)=N/N_0$), Sublinear (Eq.~\ref{eq:length_factor}). 
}
\label{tab:length_factor}
\end{table}

\subsection{Evaluation on Open-Ended Video QA}
To assess whether EcoFrame generalizes to open-ended video QA, we further evaluate it on ActivityNet-QA~\cite{yu2019activityqa} using a uniformly sampled subset of 1600 samples from the test split. ActivityNet-QA contains free-form questions over diverse human activities and evaluates the semantic correctness of generated answers. Since the dataset mainly contains shorter videos, we set the maximum frame budget to 16 frames for all methods. We use the evaluation framework integrated in lmms-eval~\cite{zhang2025lmms}, using \texttt{gpt-4o-mini} as the automatic judge for all VLM responses following official protocol.

As shown in Table~\ref{tab:open_ended_qa}, EcoFrame achieves a better overall accuracy--efficiency trade-off on ActivityNet-QA, outperforming uniform sampling and BOLT while requiring fewer final input frames on average. Notably, compared with the w/o early exit variant that always proceeds to the maximum budget, the full method achieves comparable accuracy with fewer frames and lower inference latency. This result demonstrates that entropy-gated budget scheduling remains effective in open-ended video QA.

\begin{table}[h]
\small
\begin{tabular}{lccc}
\toprule
\textbf{Method}
& \textbf{\#Frames $\downarrow$}
& \textbf{Accuracy $\uparrow$}
& \textbf{Latency $\downarrow$} \\
\midrule
Uniform
& 16.0
& 54.5 
& 0.91 \\
\midrule

BOLT
& 16.0
& 55.2
& 2.2 \\

EcoFrame-noE
& 16.0
& \textbf{55.5}
& 1.71 \\

EcoFrame
& 12.9
& 55.4
& \textbf{1.37} \\

\bottomrule
\end{tabular}
\centering
\caption{
Results on the open-ended ActivityNet-QA benchmark with LLaVA-OneVision-7B.
All methods use a maximum budget of 16 frames. EcoFrame-noE denotes the w/o early exit variant.
}
\label{tab:open_ended_qa}
\end{table}

\subsection{Additional Analysis on Scheduling Signals}

To assess the reliability of output entropy and frame-level attention on other benchmarks, we repeat the same analysis experiments of these signals on LongVideoBench and ActivityNet-QA.

\begin{figure}[h]
    \centering
    \includegraphics[width=\columnwidth]{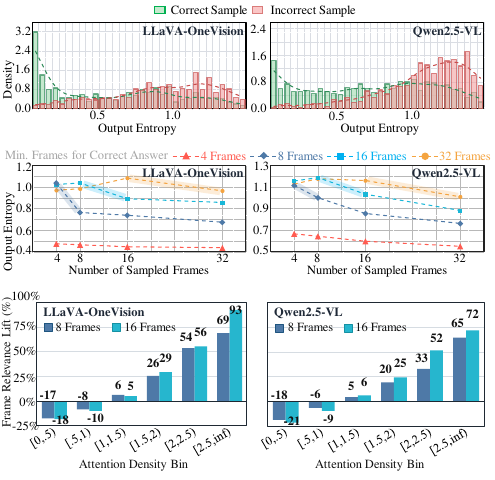}
    \caption{
        Analysis of output entropy and frame-level attention on LongVideoBench.
    }
    \label{fig:analysis_lvb}
\end{figure}

\begin{figure}[h]
    \centering
    \includegraphics[width=\columnwidth]{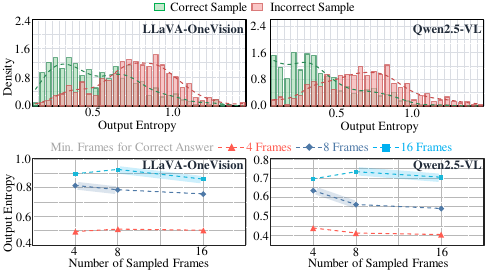}
    \caption{
        Analysis of output entropy on ActivityNet-QA.
    }
    \label{fig:analysis_act}
\end{figure}

As shown in Fig.~\ref{fig:analysis_lvb} and Fig.~\ref{fig:analysis_act}, both benchmarks exhibit patterns consistent with our main observations: lower output entropy is associated with more sufficient visual evidence, while concentrated frame-level attention more reliably identifies temporally relevant regions.

\section{Frame-Level Attention Extraction}

Previous studies have explored using VLM attention to estimate the importance of visual tokens~\cite{chen2024fastv,li2026dytok}.
FlexSelect~\cite{lu2026flexselect} finds that attention from intermediate-to-deep layers provides the most reliable importance estimates,
while Feather~\cite{endo2025feather} shows that rotary positional embeddings can introduce positional bias into attention distributions.

Inspired by these findings,
we further investigate frame-level attention as a temporal prior for long-video evidence acquisition.
Specifically, we compute attention from the query--key representations before applying positional embeddings
over a set of selected deep layers, as defined in Eq.~\ref{eq:attention} of the main paper, and average the resulting token-level scores within each frame. Since all evaluated models have 28 attention layers, we use layers 19--21 as reference layers by default.
We ablate different attention extraction schemes and choices of reference layers below.

\begin{table}[h]
\centering
\small
\setlength{\tabcolsep}{7pt}
\begin{tabular}{lcccc}
\toprule
\multirow{2}{*}{\textbf{Attn. Extraction}}
& \multicolumn{2}{c}{\textbf{LLaVA-OV}}
& \multicolumn{2}{c}{\textbf{Qwen2.5-VL}} \\
\cmidrule(lr){2-3}
\cmidrule(lr){4-5}
& \textbf{V-MME}
& \textbf{LVB}
& \textbf{V-MME}
& \textbf{LVB} \\
\midrule
w/o attention
& 60.7 & 59.2 & 63.8 & 59.3 \\
\midrule
post-RoPE (19--21)
& 61.0 & 59.7 & 64.2 & 60.1 \\
\midrule
Layers 0--3
& 60.9 & 59.8 & 64.2 & 59.4 \\
Layers 4--7
& 61.0 & 59.6 & 64.4 & 59.9 \\
Layers 8--11
& 60.8 & 59.9 & \textbf{64.5} & 59.5 \\
Layers 12--15
& 60.9 & 59.2 & 64.0 & 59.6 \\
Layers 16--18
& 60.9 & 59.5 & 64.2 & 59.8 \\
Layers 19--21
& 61.0 & \textbf{59.9} & 64.4 & \textbf{60.3} \\
Layers 22--24
& 61.0 & 59.8 & 64.1 & 60.1 \\
Layers 25--27
& \textbf{61.2} & 59.8 & 64.2 & 59.8 \\
\bottomrule
\end{tabular}
\caption{
Ablation of frame-level attention extraction.
We evaluate attention computed post-RoPE and pre-RoPE attention extracted from different groups of reference layers.
We report accuracy on Video-MME and LongVideoBench using LLaVA-Onevision-7B and Qwen2.5-VL-7B.
}
\label{tab:attention_extraction}
\end{table}

As shown in Table~\ref{tab:attention_extraction}, pre-RoPE attention extracted from certain deep layers such as 19--21 provides an overall more reliable temporal prior, achieving more consistent performance across datasets and models while consistently improving over the w/o attention variant.
\section{Limitations and Discussion}

\paragraph{Scenario-Dependent Compute Configuration.}
Although EcoFrame enables adaptive evidence scheduling for different queries, the entropy thresholds and maximum frame budget remain globally configured.
The most preferable operating point may vary across models, video domains, and deployment requirements.
Future work could adapt these compute configurations automatically through lightweight calibration or cost-aware scheduling policies.

\paragraph{Needle-in-a-Haystack Evidence.}
Extremely short evidence segments may remain challenging when they occupy only a very small fraction of a long video.
Attention-guided candidate proposal mitigates this issue by preserving global coverage when the attention distribution is diffuse, but extremely brief events may still be missed with limited candidate frames. Future work could incorporate complementary temporal cues, such as event boundaries or motion saliency, to enable more precise evidence search.


\end{document}